\documentclass[floatfix,rmp,twocolumn,twoside]{revtex4}
\usepackage{graphicx}

\bibpunct{[}{]}{,}{n}{}{,}

\usepackage{amsmath}
\usepackage[english]{babel}
\usepackage{verbatim}
\usepackage{graphicx}
\usepackage{alltt}
\usepackage[small]{caption}
\usepackage[english]{babel}

\makeatletter
\g@addto@macro\@verbatim\small
\makeatother

\begin{document}

\title{A Model-Driven Probabilistic Parser Generator}
\author{Luis~Quesada, Fernando~Berzal, and Francisco J.~Cortijo\\
  Department of Computer Science and Artificial Intelligence, CITIC, University of Granada, \\
  Granada 18071, Spain \\
  \textit{lquesada@decsai.ugr.es, fberzal@decsai.ugr.es, cb@decsai.ugr.es}
  }

\begin{abstract}
Existing probabilistic scanners and parsers impose hard constraints on the way lexical and syntactic ambiguities can be resolved.
Furthermore, traditional grammar-based parsing tools are limited in the mechanisms they allow for taking context into account.
In this paper, we propose a model-driven tool that allows for statistical language models with arbitrary probability estimators.
Our work on model-driven probabilistic parsing is built on top of ModelCC, a model-based parser generator, and enables the probabilistic interpretation and resolution of anaphoric, cataphoric, and recursive references in the disambiguation of abstract syntax graphs.
In order to prove the expression power of ModelCC, we describe the design of a general-purpose natural language parser.
\end{abstract}

\maketitle
\section{Introduction}

Natural languages suffer from lexical ambiguities and syntactic ambiguities. Lexical ambiguities \cite{Nawrocki1991} occur when a lexeme has several meanings \cite{Jurafsky2009}. Syntactic ambiguities occur when a token sequence can be generated using more than one parse tree \cite{Aho2006}.

A common approach to disambiguation consists of performing probabilistic scanning (i.e. probabilistic lexical analysis) and probabilistic parsing (i.e. probabilistic syntactic analysis), which assign a probability to each possible parse tree. However, existing techniques for probabilistic scanning and parsing present several drawbacks: probabilistic scanners may produce incorrect sequences of tokens due to wrong guesses or to occurrences of words that are not in the lexicon, and probabilistic parsers cannot consider relevant context information such as resolved references between language elements.

Model-based language specification techniques \cite{Kleppe2007} decouple language design from language processing. ModelCC \cite{Quesada2011c,Quesada2012d} is a model-based parser generator that includes support for dealing with references between language elements and, thus, instead of returning mere abstract syntax trees, ModelCC is able to obtain abstract syntax graphs and consider lexical and syntactic ambiguities.

In this paper, we explain how ModelCC supports probabilistic language models.
Section \ref{sec:background} provides an introduction to probabilistic parsing techniques and to the model-based language specification techniques employed by the ModelCC parser generator.
Section \ref{sec:probabilitysupport} explains the probabilistic model support in ModelCC.
Section \ref{sec:exampleofprobabilisticmodel} presents a case study that illustrates.
Finally, Section \ref{sec:conclusionsandfuturework} presents our conclusions and pointers for future work.

\section{Background} \label{sec:background}

In this section, we provide an analysis of the state of the art on probabilistic parsing and on model-based language specification.

\subsection{Probabilistic Parsing} \label{sec:parsing}

There are many approaches to part-of-speech tagging using probabilistic scanners and for language disambiguation using probabilistic parsers.

Probabilistic scanners based on Markov-like models \cite{Markov1971} consider the existence of implicit relationships between words, symbols or characters found close in sequences, and irrevocably guess the type of a lexeme based on the preceding ones. When using such techniques, a single wrong guess renders the whole parsing procedure irremediably erroneous, as no correct parse tree that uses a wrong token can be found.

Probabilistic scanners based on lexicons \cite{Jurafsky2009} assign probabilities to a lexeme belonging to different word classes from the statistical analysis of lexicons. Scanning a lexeme that belongs to a particular word class but never belonged to that class in the training lexicon provides wrong scanning results, which, in turn, render the whole parsing procedure useless.

Probabilistic parsers \cite{Ney1991} compute the probability of different parse trees by considering token probabilities and grammar production probabilities, which are empirically obtained from the analysis of linguistic corpora. The probability of a symbol is defined as the product of the probability of the grammar rule that produced the symbol and the probabilities of all the symbols involved in the application of that rule. The probability of a parse tree is that of its root symbol. These techniques do not take context into account.

Probabilistic lexicalized parsers \cite{Collins2003,Charniak1997} associate lexical heads and head tags to the grammar symbols. Grammar rules are then decomposed and rewritten to include the different combinations of symbols, lexical heads, and head tags. Different probabilities can be associated to each of the new rules. When using this technique, the grammar significantly expands and a more extensive analysis of linguistic corpora is needed to produce accurate results.
It should be noted that this technique is not able to consider relevant context information such as resolved references between language elements.

Conventional probabilistic scanners and parsers do not allow the use of arbitrary probability estimators or statistical models that take advantage of more context information.

\begin{figure}[tb]
\begin{minipage}[tb]{\linewidth}
\centering
\includegraphics[scale=0.215]{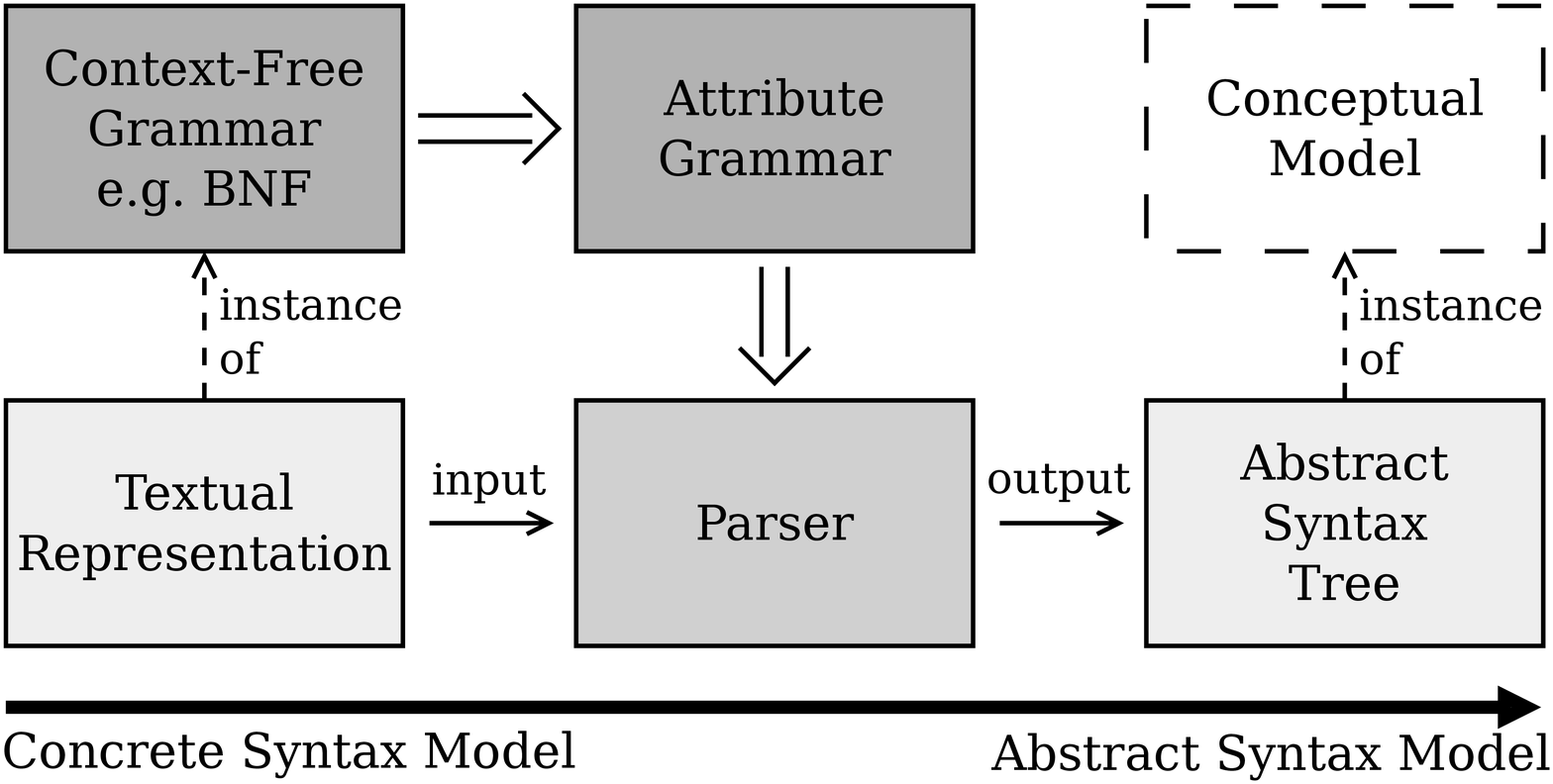}
\caption{Traditional language processing.} \label{fig:traditional}

\includegraphics[scale=0.215]{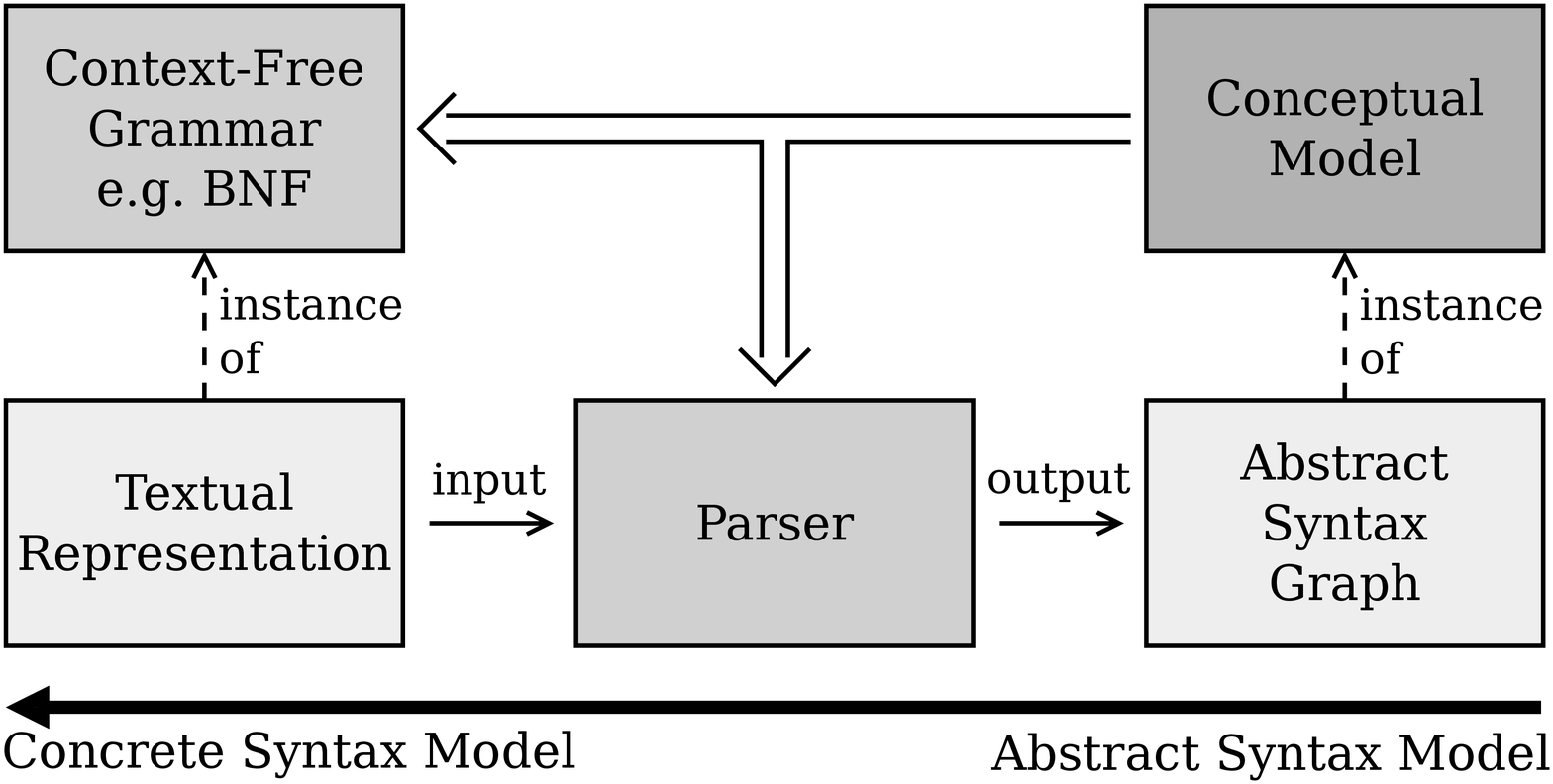}
\caption{Model-based language processing.} \label{fig:ModelCC}
\end{minipage}
\end{figure}

\subsection{Model-Based Language Specification} \label{sec:modelbased}

In its most general sense, a model is anything used in any way to represent something else. In such sense, a grammar is a model of the language it defines. The idea behind model-based language specification is that, starting from a single abstract syntax model (ASM) that represents the core concepts in a language, language designers can develop one or several concrete syntax models (CSMs). These CSMs can suit the specific needs of the desired textual or graphical representation for language sentences. The ASM-CSM mapping can be performed, for instance, by annotating the abstract syntax model with the constraints needed to transform the elements in the abstract syntax into their concrete representation.

A diagram summarizing the traditional language design process is shown in Figure \ref{fig:traditional}, whereas the corresponding diagram for the model-based approach is shown in Figure \ref{fig:ModelCC}. It should be noted that ASMs represent non-tree structures whenever language elements can refer to other language elements, hence the use of the `abstract syntax graph' term.

ModelCC \cite{Quesada2011c,Quesada2012d} is a parser generator that supports a model-based approach to the design of language processing systems.
Its starting ASM is created by defining classes that represent language elements and establishing relationships among those elements. Once the ASM is created, constraints can be imposed over language elements and their relationships as metadata annotations \cite{Fowler2002} in order to produce the desired ASM-CSM mappings.

Although probabilistic language processing techniques and model-based language specification have been extensively studied, to the best of our knowledge, there are no techniques that allow model-driven probabilistic parsing. In the next section, we explain ModelCC's support for probabilistic language models.

\section{Probabilistic Parsing in ModelCC} \label{sec:probabilitysupport}

ModelCC effectively combines model-based language specification with probabilistic parsing by allowing the specification of arbitrary probabilistic language models.

Subsection \ref{sec:custom} introduces ModelCC support for probabilistic language models and presents ModelCC's \emph{@Probability} annotation.
Subsection \ref{sec:context} discusses the use of contextual information in ModelCC probabilistic parsers.
Subsection \ref{sec:probability} explains how symbol probabilities are computed.

\subsection{Probabilistic Language Models} \label{sec:custom}

ModelCC's \emph{@Probability} annotation allows the specification of probability values for language elements and language element members. Probability values can be specified for syntactic elements of the languages and for lexical components, in which case it should be noted that the lexical analyzer behaves as a part-of-speech tagger in natural language processing.

Such probability values can be specified using three alternatives: a probability value as a real number between $0$ and $1$, a frequency as an integer number, or a custom probability evaluator that computes the probability value from the analysis of the language element and its context.

Since ModelCC supports lexical and syntactic ambiguities, and the combination of language models, one of the main novelties of ModelCC with respect to existing techniques is that it allows the modular specification of probabilistic languages, that is, it is able to produce parsers from composite language specifications even when some of the language elements overlap or conflict.

ModelCC also supports alternative models for the representation of uncertainty (e.g. possibilistic models, models based on Dempster-Shafer theory, or any other soft computing models), provided that an evaluation operator for language element instances is provided and an evaluation operator for the application of grammar rules is provided.
Optionally, a casting operator that translates the estimated value in one model into a value valid for a different kind of model allows the specification of modular languages even when different mechanisms for representing natural language ambiguities are employed for different parts of the language model.

\subsection{Context Information} \label{sec:context}

ModelCC provides context information that custom probability evaluators and constraints can take into account when processing a language element.

The context information includes the current syntax graph and the parse graph symbol corresponding to the language element being evaluated.
Also, if the language element instance is a reference, the context information also includes the referenced language element instance, its corresponding parse graph symbol, and the context graph, which is the smallest graph that contains both the reference and the referenced object.

It should be noted that, from this information, it is possible to deduce traditional metrics such as the distance between the reference and the referenced object in the input or in the syntax graph and whether the reference is anaphoric, cataphoric, or recursive.

However, in contrast to existing probabilistic parsing techniques, ModelCC also allows the specification of complex syntactic constraints, semantic constraints, and probability evaluators that use extensive context information such as resolved references between language elements. 

\subsection{Probability Evaluation} \label{sec:probability}

The probability of a particular parse graph $G$ for a sentence $w_{1:m}$ of length $m$ is defined as the product of the probabilities associated to the $n$ instances of language elements $E_i$ in the parse graph $G$:

\begin{equation}
P(G|w_{1:m}) = \prod_{i=1}^n P(E_i|w_{s_i:e_i})
\end{equation}
 
Given a language element $E$ that represents a part-of-speech tag and a word $w$, the lexical analyzer acts as a POS tagger and provides $P(E|w)$.

Given a language element $E$ with $M_1..M_n$ members in its definition, some of which are optional, the probability $P(E|M_{1:n})$ is computed as follows. Let $OPT(E)$ be the set of optional elements for $E$. Assuming that their appearance is statistically independent, we can estimate the probability of $E$ given its observed elements $O$:
 
\begin{equation}
P(E|O_{1:k}) = P(E) \prod_{\substack{M_i \in OPT(E), \\ M_i \in O_{1:k}}} P(M_i|E) \prod_{\substack{M_j \in OPT(E),\\M_j \notin O_{1:k}}} (1- P(M_j|E))
\end{equation}

Given an ambiguous sentence $w_{1:n}$, its disambiguation is done by picking the parse graph $\hat{G}$ with the highest probability for that sentence:

\begin{figure*}[p]
\begin{minipage}[p]{\linewidth}
\centering

\includegraphics[scale=0.98]{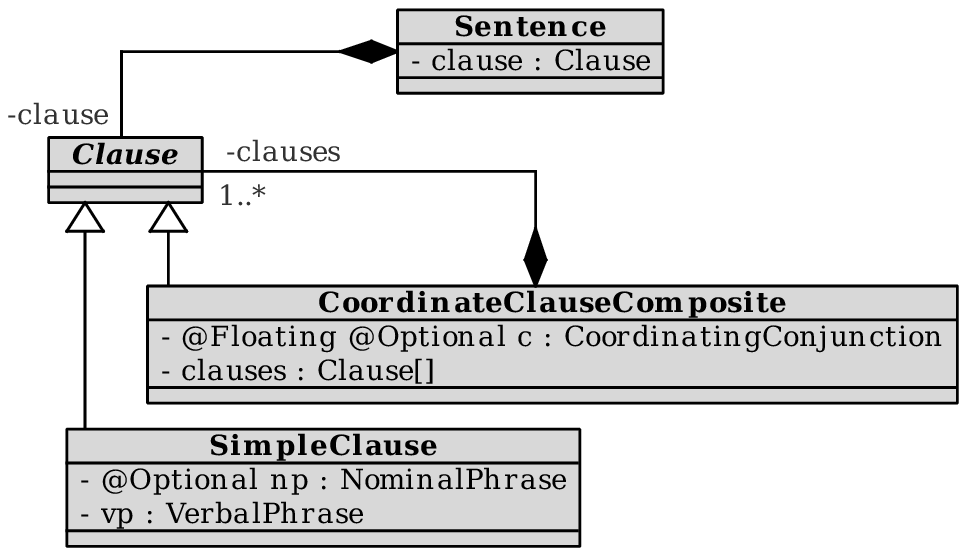}
\caption{ModelCC specification of the sentence and clause elements of our general natural language.} \label{fig:modelsencla}

\vspace{17mm}
\includegraphics[scale=0.98]{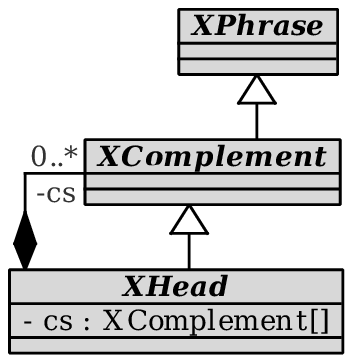}
\caption{ModelCC specification of the phrase, head, and complement elements of our general natural language.} \label{fig:modelphheco}

\vspace{17mm}
\includegraphics[scale=0.98]{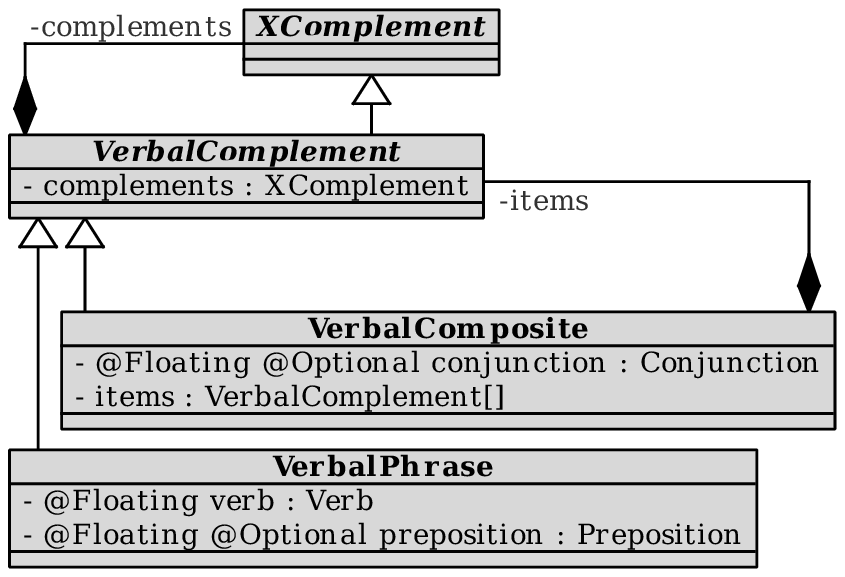}
\caption{ModelCC specification of the verbal complement language elements in our general natural language.} \label{fig:modelverb}
\end{minipage}
\end{figure*}

We now proceed to present an example of natural language specification using ModelCC.

\begin{figure*}[p]
\begin{minipage}[p]{\linewidth}
\centering

\includegraphics[scale=0.98]{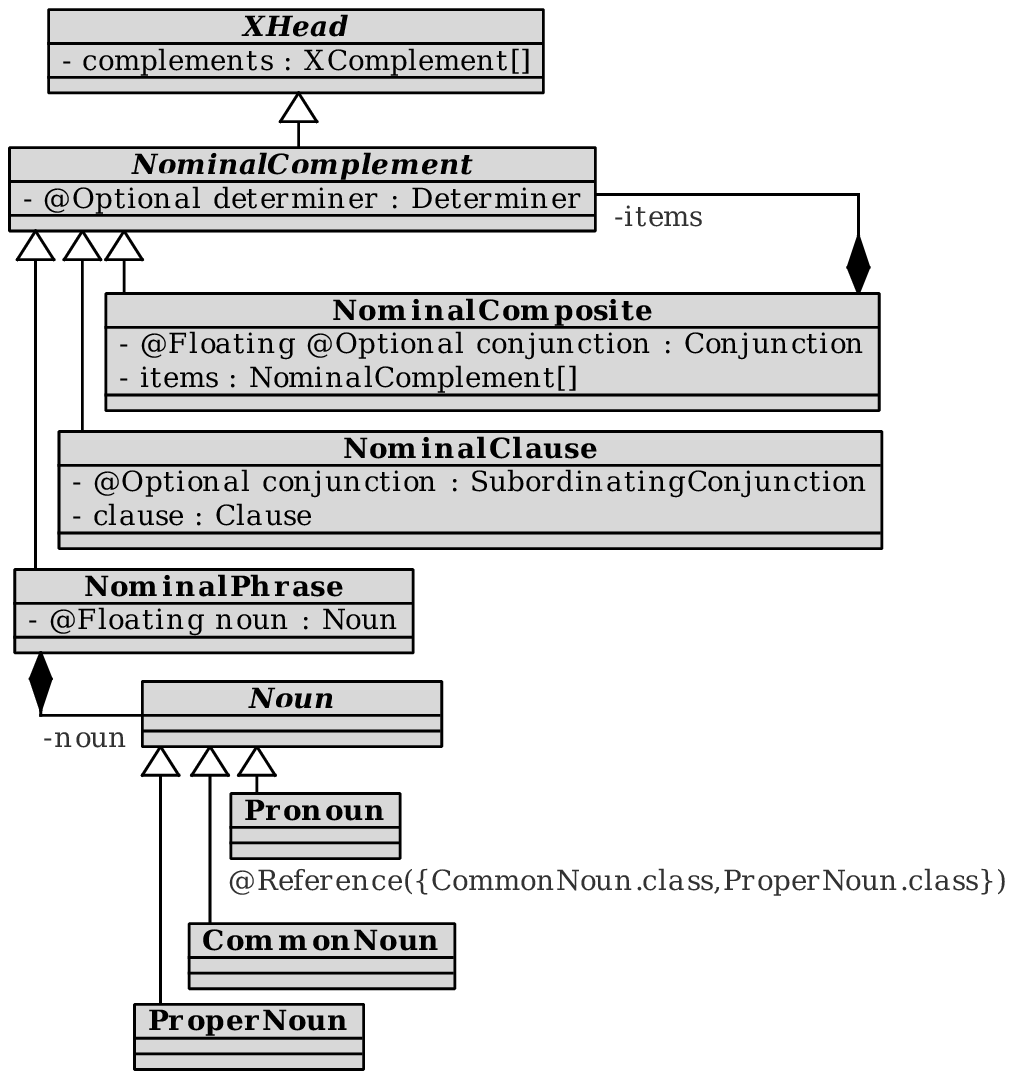}
\caption{ModelCC specification of the nominal complement language elements in our general natural language.} \label{fig:modelnom}

\vspace{17mm}
\includegraphics[scale=0.98]{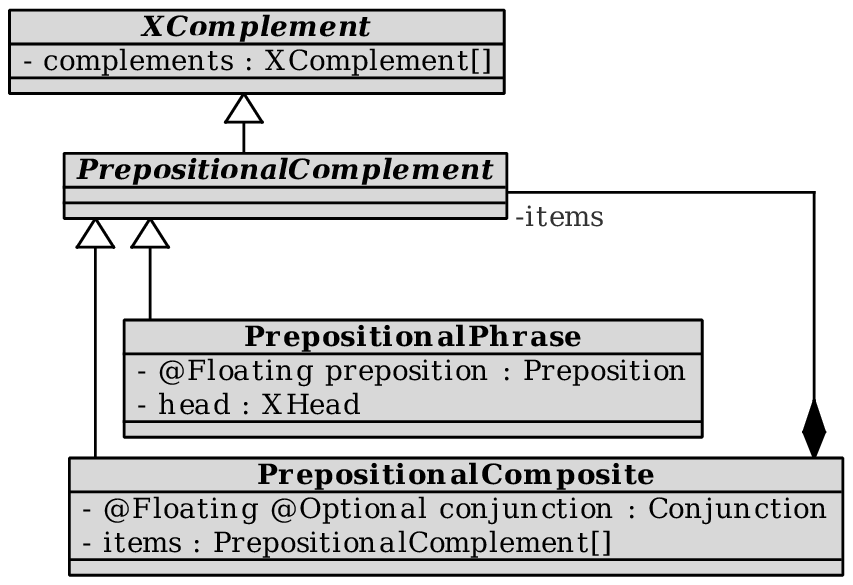}
\caption{ModelCC specification of prepositional complement language elements in our general natural language.} \label{fig:modelprep}

\end{minipage}
\end{figure*}
\begin{equation}
\hat{G}(w_{1:m}) = \arg\max_{\hspace{-5mm}G}\{P(G|w_{1:m})\}
\end{equation}

\section{Model-Based Specification of Natural Languages} \label{sec:exampleofprobabilisticmodel}

In this section, we present a model-based specification for a probabilistic natural language parser.
Subsection \ref{sec:langdesc} outlines the general natural language features.
Subsection \ref{sec:modimp} provides the ModelCC ASM specification of the general natural language.
Subsection \ref{sec:instant} explains how the general natural language can be instantiated.
Subsection \ref{sec:eng} presents a sample English language parser.

\subsection{Natural Language Description} \label{sec:langdesc}

Our general language model supports Chomsky's X-bar theory \cite{Chomsky1970}, which claims that certain human languages share structural similarities.

In our model, a sentence consists of a clause (i.e. a complete proposition), a clause can be either a simple clause or the coordinate clause composite that creates a compound sentence, a simple clause consists of an optional nominal phrase and a verbal phrase, and a coordinate clause composite consists of a set of clauses and an optional floating coordinating conjunction.

In our general natural language model, a complement is a phrase used to complete a predicate construction, and a head is a complement that plays the same grammatical role as the whole predicate construction.

Our general natural language model supports nominal, verbal, adverbial, adjectival, and prepositional complements.

Nominal complements comprise nominal phrases, nominal composites, and nominal clauses.
A nominal phrase consists of an optional determiner, a noun, and an optional set of complements. 
A nominal composite consists of an optional determiner, a set of nominal complements and an optional floating conjunction.
A nominal clause consists of an optional determiner, an optional subordinating conjunction and a subordinate clause.
Nouns comprise common nouns, proper nouns, and pronouns. Pronouns, in turn, reference nouns and proper nouns.

Verbal complements comprise verbal phrases and verbal composites.
A verbal phrase consists of a set of floating verbs and an optional floating preposition.
A verbal composite consists of a set of verbal complements and an optional floating conjunction.

Adverbial complements comprise adverbial phrases, adverbial composites, and adverbial clauses.
An adverbial phrase consists of an adverb.
An adverbial composite consists of a set of adverbial complements and an optional floating conjunction.
An adverbial clause consists of an optional subordinating conjunction and a subordinate clause.

Adjectival complements comprise adjectival composites and adjectival clauses.
An adjectival composite consists of a set of adjectival complements and an optional floating conjunction.
An adjectival clause consists of an \clearpage \noindent optional subordinating conjunction and a subordinate clause.

Prepositional complements comprise prepositional phrases and prepositional composite.
A prepositional phrase consists of a floating preposition and a head.
A prepositional composite consists of a set of prepositional complements and an optional floating conjunction.

It should be noted that this general natural language embraces Romance languages such as Spanish, Portuguese, French, and Italian, as well as Germanic languages such as English and German.

\subsection{ModelCC Specification of the ASM for Natural Languages} \label{sec:modimp}

In order to implement our general natural language parser using ModelCC, we have to provide a specification of the language ASM. This specification is provided as a set of UML diagrams, as illustrated in Figures \ref{fig:modelsencla}, \ref{fig:modelphheco}, \ref{fig:modelverb}, \ref{fig:modelnom}, and \ref{fig:modelprep}. Adjectival complements and adverbial complements can be specified as nominal complements in Figure \ref{fig:modelnom}. As it can be observed from the figures, the model-based specification of the general natural language is easily obtained from the language description requirements.

As the specified model is an abstract syntax model, it does no correspond to any particular model. The ASM is more like the Mentalese language postulated by the Language Of Thought Hypothesis \cite{sep-language-thought}.
In the next subsection, we explain how different fully-functional natural language parsers can be instantiated from this model by defining additional language-specific constraints.

\subsection{Specification of the Natural Language CSMs} \label{sec:instant}

In order to implement a parser for a particular natural language, the ASM-CSM mapping has to be specified.
A pattern matcher is assigned to each lexical component of the language model.
For this purpose, ModelCC's \emph{@Pattern} annotation allows the specification of custom pattern matchers that can consist of regular expressions, dictionary lookups, or any suitable heuristics. Such pattern matchers can easily be induced from the analysis of lexicons.

ModelCC supports lexical ambiguities apart from syntactic ambiguities, so the specified pattern matchers can produce different, and even overlapping, sets of tokens from the analysis of the input string.

After specifying the pattern matchers for the lexical components of the language, language-specific constraints are assigned to syntactic components of the language model.
For this purpose, ModelCC's \emph{@Constraint} annotation allows the specification of methods that evaluate whether a language element instance is valid or not.
These constraints can be automatically induced from the analysis of linguistic corpora and, as explained in Subsection \ref{sec:context}, these constraints can take into account extensive context information, which can even include resolved references between language elements.

Finally, in order to produce a probabilistic parser, probability evaluators are assigned to the different language constructions.
For this purpose, ModelCC's \emph{@Probability} annotation allows the specification of the probability evaluation for language elements.
These probabilities can also be estimated from the analysis of linguistic corpora, although heuristics could also be used.

\subsection{An Example: Parsing an English Sentence} \label{sec:eng}

We have implemented an English parser by specifying an ASM-CSM mapping from the general language ASM.

We have defined pattern matchers that query wiktionary.org to perform the lexical analysis.
We have approximated probability values derived from the analysis of the Google n-gram datasets to different lexemes and constructions.

\begin{figure*}[tb]
\begin{minipage}[tb]{\linewidth}
\centering
\includegraphics[scale=0.5]{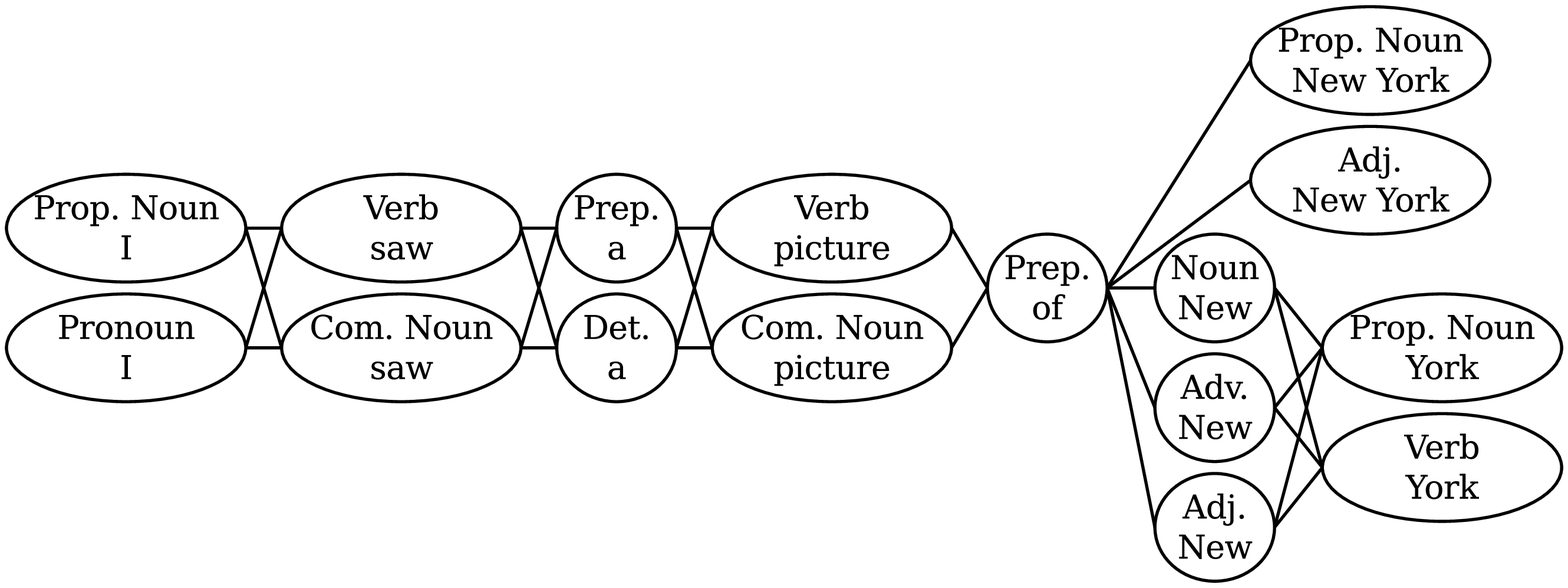}
\caption{Lexical analysis graph for the sentence ``I saw a picture of New York''.} \label{fig:york}

\vspace{7mm}

\includegraphics[scale=0.5]{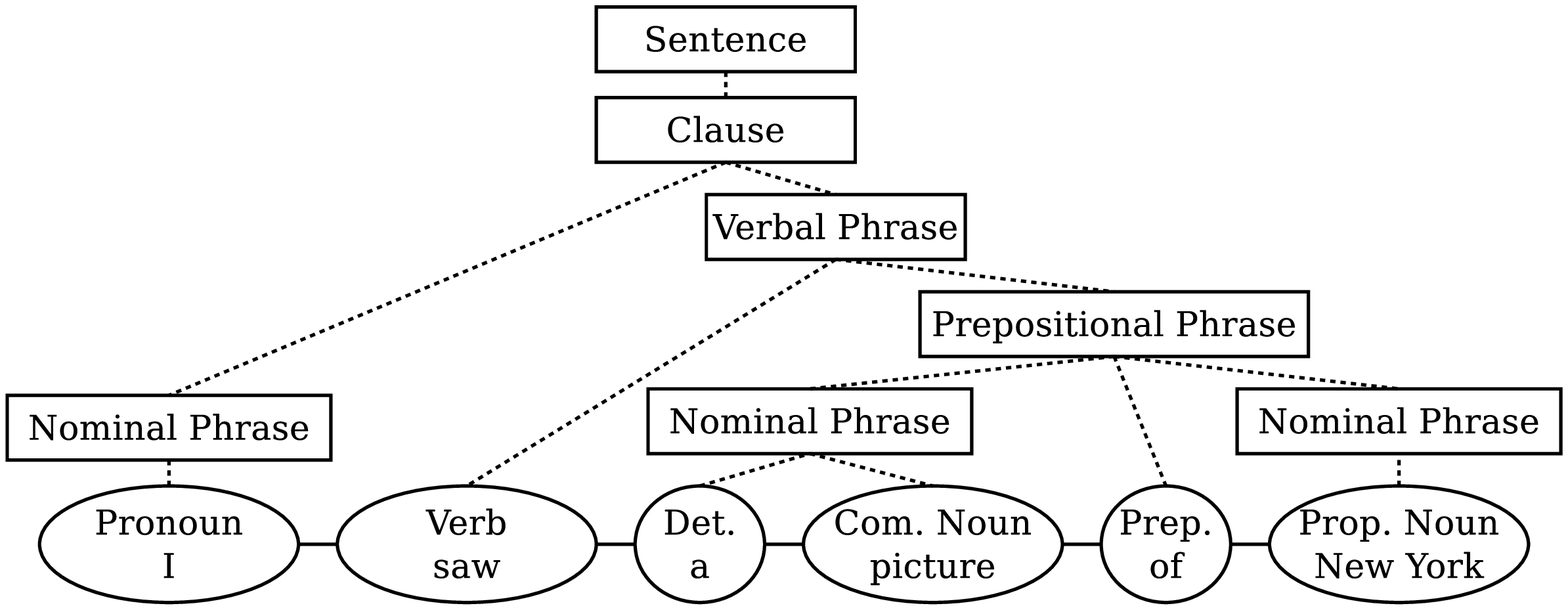}
\caption{Correct parse graph for the sentence ``I saw a picture of New York''.} \label{fig:york2}

\end{minipage}
\end{figure*}

As an example, we have parsed the sentence ``I saw a picture of New York''.
The lexical analysis graph for this sentence represents 128 valid token sequences and is shown in Figure \ref{fig:york}.
A set of valid parse graphs can be obtained from this lexical analysis graph and Figure \ref{fig:york2} shows the correct parse tree.

\section{Conclusions and Future Work} \label{sec:conclusionsandfuturework}

Natural languages suffer from ambiguities. A common approach to disambiguation consists of performing probabilistic scanning and probabilistic parsing.
Such techniques present several drawbacks: wrong sequences of tokens may be produced, and only small amounts of context information are used.

We have described ModelCC's support for probabilistic language models.
ModelCC is a model-based parser generator that supports lexical ambiguities, syntactic ambiguities, and reference resolution.

Also, we have demonstrated the application of ModelCC to probabilistic parsing by providing a model-based specification of a general natural language, providing an English-language instantiation of it.

We plan to do research on the automatic induction of probabilistic language models, syntactic constraints, and semantic constraints from linguistic corpora. We also plan to do research on the application of alternative models for the representation of uncertainty to natural language parsing.

\bibliographystyle{plain}
\bibliography{doc}

\end{document}